\documentclass[pdflatex,sn-basic, iicol]{sn-jnl}

\usepackage{upgreek}
\usepackage{xcolor}
\usepackage[]{graphicx}

\jyear{2021}%

\theoremstyle{thmstyleone}%
\theoremstyle{thmstyletwo}%

\theoremstyle{thmstylethree}%

\newcommand{\repeatfoot}{\textsuperscript{\thefootnote}}

\begin{document}

\title[Article Title]{A Stochastic Variance-Reduced Coordinate Descent Algorithm for Learning Sparse Bayesian Network from Discrete High-Dimensional Data}


\author*[1]{\fnm{Nazanin} \sur{Shajoonnezhad}}\email{n.shajoonnezhad@email.kntu.ac.ir}

\author[2]{\fnm{Amin} \sur{Nikanjam}}\email{amin.nikanjam@polymtl.ca}

\affil[1]{\orgname{K. N. Toosi University of Technology}, \orgaddress{\city{Tehran},  \country{Iran}}}

\affil[2]{ \orgname{Polytechnique Montréal}, \orgaddress{\state{Quebec}, \country{Canada}}}


\abstract{This paper addresses the problem of learning a sparse structure Bayesian network from high-dimensional discrete data. Compared to continuous Bayesian networks, learning a discrete Bayesian network is a challenging problem due to the large parameter space. Although many approaches have been developed for learning continuous Bayesian networks, few approaches have been proposed for the discrete ones. In this paper, we address learning Bayesian networks as an optimization problem and propose a score function which guarantees the learnt structure to be a sparse directed acyclic graph. Besides, we implement a block-wised stochastic coordinate descent algorithm to optimize the score function. Specifically, we use a variance reducing method in our optimization algorithm to make the algorithm work efficiently for high-dimensional data. The proposed approach is applied to synthetic data from well-known benchmark networks. The quality, scalability, and robustness of the constructed network are measured. Compared to some competitive approaches, the results reveal that our algorithm outperforms some of the well-known proposed methods.}




\maketitle

\section{Introduction}\label{sec1}
Bayesian Networks (BNs) are probabilistic graphical models that have been widely used for representing dependencies and independencies among variables of a problem \citep{ref1, ref2}. One can consider a BN as a compact representation of a joint probability distribution function \citep{ref3}. A BN consists of a structure which is a Directed Acyclic Graph (DAG), and a parameter set which represents the quantitative information about dependencies among a set of variables. These networks have been widely used in machine vision \citep{ref4}, bioinformatics \citep{ref5}, data fusion \citep{ref6}, and decision support systems \citep{ref7}. Although many approaches have been developed for learning a continuous BN, there is a lack for the discrete one. The reason is that learning a discrete BN is a challenging problem due to the large parameter space and the difficulty in searching for an efficient structure. However, the problem becomes more challenging when we need to learn a sparse BN from a high-dimensional discrete data. There is great demand for sparse structures in a broad collection of problems from brain sciences \citep{ref29} to biological sciences \citep{ref28}. For example, identifying functional brain networks \citep{zhu2021}, or modeling the interacting patterns between genes from microarray gene expression data \citep{kourour} represents high-dimensional data, in which the number of samples is equal or less than the number of variables \citep{ref56}. In addition, many real-world networks, such as gene association networks and brain connectivity networks are sparse \citep{ref53}. Hence, accurately learning a sparse structure from such datasets is of great importance.

Most methods that have been proposed for BN structure learning fall into three categories: 1) constraint-based methods \citep{ref8} such as Principal Component (PC) \citep{ref13}, Max-Min Parents and Children (MMPC) \citep{ref14}, and Fast Causal Inference (FCI) \citep{ref15}, 2) score-based methods \citep{ref9,ref10}, and 3) hybrid methods \citep{ref11,ref12} such as MAX-Min Hill-Climbing (MMHC) \citep{ref14}. With growing tendency toward sparse modeling, score-based methods have attracted more attention since their capability of applying constraints to the score functions. These methods assign a score to each structure and then search for the structure with the best score. Different score functions such as Bayesian Dirichlet (BD) metric, Bayesian Information Criterion (BIC), Minimum Description Length (MDL), and entropy-based metrics are used in score-based structure learning. After assigning scores, a search algorithm is used to find the optimal structure (with the optimal score). Various algorithms have been employed for the search step in score-based structure learning methods, like Hill-Climbing (HC), k2 \citep{ref17} and Monte Carlo methods \citep{ref18,ref19}. However, finding an exact structure is an NP-Hard problem \citep{ref20,ref21}, hence, heuristic search techniques \citep{ref22}, genetic algorithms \citep{ref23,ref24} and simulated annealing \citep{ref25} were employed to address this problem.
  
By adding a penalty term to the score function, conditions such as sparsity and DAG property of BN can be modeled in the score function. By optimizing the new score function, we can find the best sparse structure. Sparse Candidate (SC) \citep{ref30}, and Sparse Bayesian Network (SBN) \citep{ref1}, which are proposed for Gaussian data are examples of this approach. Zheng et al. \citep{ref31} have proposed NOTEARS, a score-based method that formulates the structure learning of linear Structural Equation Models (SEMs) as a continuous optimization problem using a smooth characterization of acyclicity. Subsequent works such as DAG-GNN \citep{ref54} and GraN-DAG \citep{ref55} have extended NOTEARS to handle nonlinear cases. With smooth score functions, these methods utilize gradient-based optimization to learn DAGs by estimating some weighted graph adjacency matrices. NOTEARS and DAG-GNN assume specific forms of SEMs where weighted adjacency matrices naturally exist; their performance usually degrades when the data model does not follow these forms. Recently, Yu et al. have proposed DAG-NoCurl \citep{ref32}, in which they showed that the set of weighted adjacency matrices of DAGs are equivalent to the set of weighted gradients of graph potential functions, and one may perform structure learning by searching in this equivalent set of DAG. This approach solves the optimization problem with a two-step procedure: 1) finding an initial cyclic solution to the optimization problem, and 2) employing the Hodge decomposition of graphs and learning an acyclic graph by projecting the cyclic graph to the gradient of a potential function. \cite{ref33} used a maximum likelihood function with an L1 penalty term as score function. Afterward, this algorithm has been expanded to use a concave penalty term \citep{ref34}.

While most of the research about BNs is concentrated on the continuous domain, learning discrete BNs is challenging because of large parameter space and difficulty in searching for a sparse structure. Recently, a Block-wised Coordinate Descent (BCD) method has been proposed for learning discrete BN’s structure \citep{ref35}. The proposed algorithm is a deterministic method which has employed a high-order optimization method based on Newton-Raphson method, to optimize the score function. BCD algorithm introduces an L1 regularized likelihood score function and a constraint has been applied to the score function to guarantee the sparsity. The algorithm achieves competitive results, but it needs further efforts to ensure the acyclicity constraint, which results in some potential structures not being checked during the learning process. In addition, although high-order optimizations attract widespread attention, the operation and storage of the inverse matrix of the Hessian matrix make some challenges for them \citep{ref51}.

The Gradient Descent (GD) based methods are simple to implement. Besides this, the solution in this type of optimization is globally optimal when the objective function is convex. However, the cost is hard to accept when dealing with large-scale data: if the dimension of data is $N \times D$, in which $N$ is the number of data and $D$ is the number of features, the computation complexity for each iteration will be $O(ND)$. To overcome this problem, the Stochastic Gradient Descent (SGD) method emerges \citep{ref51}. While stochastic methods have shown better performance in high-dimensional problems, the gradient direction in SGD oscillates because of additional noise introduced by random selection, and the search process becomes blind in the solution space. As a solution, we have to decrease the step-size to zero step-by-step, which consequently leads to the slower convergence pace.

In this paper, we have proposed a stochastic score-based method called Stochastic Variance Reduction Coordinate Descent (SVRCD) for discrete BN sparse structure learning. We have proposed a new penalized Log-Likelihood (LL) score function and developed an optimization algorithm accordingly. To achieve a sparse directed acyclic structure, we have added two penalties to the likelihood term to define the new score function, one for controlling sparsity, and another for achieving a directed acyclic structure \citep{ref9}. The optimization algorithm uses a BCD method in which each coordinate is optimized using SGD. We have utilized a reducing variance method \citep{ref36} in SGD to accelerate the optimization process. The efficiency, scalability, and robustness of the algorithm are measured. The results show that SVRCD outperforms some of the well-known algorithms in BN structure learning. We summarize the contributions of this study as follows:
\begin{itemize}
    \item We have presented a novel score function for discrete Bayesian networks which consists of two penalty-terms, one for ensuring the sparsity and another one for converging to a DAG structure.
    \item We have solved the learning task as an optimization problem using a stochastic method instead of deterministic ones: a BCD algorithm which utilizes SGD as the optimizer.
    \item We have reduced the variances of estimations using a variance reduction method to improve the convergence time in our algorithm.
    \item We make the code (the proposed algorithm and dataset generator) used in this study publicly available online\footnote{\url{https://github.com/nshajoon/SVRCD-Algorithm.git}} for other researchers/practitioners to replicate our results or build on our work.
\end{itemize}

The rest of the paper is organized as follows: Section 2 briefly reviewed the BN and required definitions. Section 3 presents our novel score function definition and its components. Section 4 presents the SVRCD algorithm. Section 5 reports the result of the proposed algorithm for simulated networks. Concluding remarks are presented in Section 6.

\section{Preliminaries}
A BN is a graphical representation for a joint probability distribution function. This graphical model consists of a parameter set and a structure. The structure is a DAG $G=(V,E)$ that $V$ is the set of nodes, and $E$ is the set of edges $(E={i \rightarrow j\mid i,j \in V})$ , where $i$ and $j$ are respectively parent and child. Given $G$, joint probability distribution function is represented as:

\begin{equation}\label{eq1}
      P(X_{1},...,X_{p})=\prod_{i=1}^{p} P(X_{i} \mid \Uppi_i^G)      
\end{equation}
where $X_{i}$ is a random variable for $V_{i}$, and $\Uppi_i^G$  is parent set of $X_{i}$. Because of the big parameter space and the difficulty in finding a sparse structure, BN structure learning is a challenge. The score-based approach is a popular structure learning method. In this approach, the score function assigns a numeric value to each structure showing how the data fits the structure. Given dataset $D$ and graph $G$, the score is represented as:
\begin{equation}\label{eq2}
    Score(G\mid D)=P(G\mid D)=\frac{P(D\mid G)P(G)}{P(D)}
\end{equation}

 We consider the equal probability for all structures, and therefore, just the numerator of the equation \ref{eq2} need to be optimized. If the structure score is a summation of all variables scores, the score function is called decomposable. Given a variable and its parents, the variable score is:
\begin{equation}\label{eq3}
    Score(G\mid D)=\sum_{i=1}^{n}Score(X_{i}\mid \Uppi_i,D)
\end{equation}

Most score functions have a Penalized Log-Likelihood (PLL) form. We can formulate the constraints by adding a penalty term to LL function. Given $(D,G)$ an LL function is defined as:
\begin{equation} \label{eq4}
 \begin{split}
    LL(D\mid G) & =\sum_{h=1}^{n}\log P(D_h\mid G)\\
    & = \sum_{i=1}^{p}\sum_{h=1}^{n}\log P(D_{ih}\mid \Uppi_{ih})
 \end{split}
\end{equation}
where $D_{ih}$ is the value of $X_i$ in data row h, and $\Uppi_{ih}$  is the value of $X_i$’s parents in data row h. A PLL function is defined as:
\begin{equation}\label{eq5}
    PLL(G\mid D)=LL(D\mid G)-\sum_{i=1}^{n}Penalty(X_i,G,D)
\end{equation}
where $\sum_{i=1}^{n}Penalty(X_i,G,D)$  is the penalty term that has been added to the score function.

There are different optimization methods for optimizing the score function. A Coordinate Descent (CD) method is an iterative optimization method which in each iteration optimizes the function given one component. This method decomposes the problem to some reduced dimension sub-problems \citep{ref38}. A BCD algorithm is a CD algorithm that optimizes the score function using a subset of all components in each iteration.

\begin{algorithm}[t]
 \caption{CD Algorithm}\label{algo1}
\begin{algorithmic}[1]
 \Require $\gamma > 0$
 \State $k\leftarrow 0$, $x^0 \in \mathbb{R}$
 \While{termination test not satisfied}
    \State choose a random $i \in \{1...n\}$
    \State $x^{k+1}\leftarrow x^k-\gamma(\nabla h_i(x))$
 \EndWhile
\end{algorithmic}

\end{algorithm}
 
Algorithm 1 is a CD algorithm which has two steps: choosing a component of the score function, and optimizing the score function with regard to the selected component using a gradient descent (GD) method. Using SGD method in the second step, we could have better performance in high-dimensional datasets. This method optimizes the score function using just a mini batch of data which is selected stochastically.

A convex function has just one global and no local optimum. Convex functions are very important in optimization problems, since formulating a problem as a convex function could help to find the global optimum. Commonly, a quadratic optimization method such as Newton-Raphson is used for optimizing convex functions. Quadratic methods need calculating the Hessian of the function that is impossible or too complex in most problems. Instead of quadratic methods we can use GD and SGD with an appropriate learning rate $\gamma$. GD and SGD could find global optimum in a convex function. In addition, these methods use the first derivative of a function that is simply calculable.

\section{The Proposed Score Function}\label{sec3}
In this section, Bayesian network structure learning problem has been formulated as an optimization problem by defining an appropriate score function. The proposed score function is a PLL function, with two penalty terms added in order to control structure acyclicity and sparsity simultaneously. Consider $X_i$ as a discrete variable of a BN whose domain has $n_i$ value ${1,2,…,n_i}$. $\Uppi_i^G$ is $X_i$’s parents that has $\pi_k$ value ${1,2,…,\pi_k}$ in its domain. The parameter set of a BN is:

\begin{equation}\label{eq6}
   \boldsymbol\Uptheta=\{\Uptheta_{ijk}\geq 0: \Uptheta_{ijk}=P(X_i=j\mid\Uppi_i^G=\uppi_k\}
\end{equation}
The number of parameters is:
\begin{equation}\label{eq7}
    N(\boldsymbol\Uptheta)=\sum_{i=1}^{p}n_i\prod_{j\in \Uppi_i^G}^{}n_j
\end{equation}
If $n_i=O(n)$ for all $X_i$, the number of parameters is:
\begin{equation}\label{eq8}
    N(\boldsymbol\Uptheta)=O(\sum_{i=1}^{p}n^{1+\lvert \Uppi_i^G\rvert})
\end{equation}
We use a multi-logit model \citep{ref35} to decrease the number of parameters. Using this model the likelihood function is:
\begin{equation}\label{eq9}
    P(X_i=\ell \mid \Uppi_i^G)=\frac{\exp(\textbf{x}^T\beta_{i\ell.})}{\sum_{m=1}^{n_i}\exp(\textbf{x}^T \beta_{im.})}\triangleq p_{i\ell}(\textbf{x})
\end{equation}
where we code each $X_i$ using $d_i=n_i-1$ dummy variables as a vector $\boldsymbol{X_i}=(x_{i1},x_{i2},…,x_{id_i} ) \in\{0,1\}^{d_i}$; if $X_i=l$ then $x_{il}=1$ and $x_{i./l}=0$, and if $X_i=0$ then we say $X_i$ has the reference value.\\
$\boldsymbol\beta_{i\ell.}=[\beta_{i\ell0},\boldsymbol\beta_{i\ell1},...,\boldsymbol\beta_{i\ell p}] \in \mathbb{R}^d$ is a coefficient vector for predicting how probable $X_i=l$ is. If $j \notin \Uppi_i^G$ then $\boldsymbol\beta_{i.j}=0$. We use a symmetric multi-logit model and add two constraints to the model since the model is identifiable \citep{ref39}:
\begin{equation}\label{eq10}
\begin{aligned}
    &\beta_{i10}=0,
    &\sum_{m=1}^{r_j}\boldsymbol\beta_{imj}=0,\quad \forall i,j=1,...,p
\end{aligned}
\end{equation}
The number of parameters in this model is:
\begin{equation}\label{eq11}
     N(\boldsymbol\beta)=\sum_{i=1}^{p}[(n_i-1)+n_i\sum_{j \in \Uppi_i^G}^{}d_j]
\end{equation}
Consider $n_i=O(n)$ for all $i$, then:
\begin{equation}\label{eq12}
    N(\boldsymbol\beta)=O(n^2)|E|+O(np)
\end{equation}

Multi-logit model is an appropriate estimation of the product multinomial model. In this model, the number of parameters increases linearly with the number of edges. This model is much more efficient than the product multinomial model which grows exponentially as the size of the parent set increases. 
We consider dataset $D=(d_{hi})_{n\times p}$, that has been sampled from $G$, where $d_{hi}$ is the value of $X_i$ in data row $h$. Using multi-logit model, $LL(\boldsymbol\beta)$ could be defined as:
\begin{equation}\label{eq13}
\begin{aligned}
    LL(\boldsymbol\beta) & = \sum_{i=1}^{p}\sum_{h\in O_i}^{}\log [P(d_{hi}\mid x_{h,j}, j\in \Uppi_i^G)]\\
    & =  \sum_{i=1}^{p}\sum_{h\in O_i}^{}\bigg[ \sum_{\ell =1}^{r_i}y_{hi \ell}x_h^T\boldsymbol\beta_{i\ell.}\\
    & \quad \quad \quad \quad \quad -\log {\sum_{m=1}^{r_i}\exp(x_h^T\boldsymbol\beta_{im.})}\bigg] 
\end{aligned}    
\end{equation}
where $y_{hi\ell}=I(d_{hi}=\ell)$ is index variable, and $\boldsymbol\beta_{i\ell k}=0$ for all $k \notin \Uppi_i^G$. 
The score function controls the structure sparsity using a penalty term. We use a group norm penalty term that penalizes all components of the score function at the same. Let $G$ be a graph that is learned from parameter set $\boldsymbol\beta$. The sparsity penalty term is defined as:
\begin{equation}\label{eq14}
    SparsityPenalty=\lambda_1\sum_{i=1}^{p}\sum_{j=1}^{p}{\lvert}{\rvert}\boldsymbol\beta_{i.j}{\lvert}{\rvert}_2
\end{equation}
We use a different penalty term that causes the structure to not have any directed cycles. Let us consider $A^G$ and $PATH^G \in \{0,1\}^{p\times p}$  are adjacency and path matrices, respectively. Path matrix $PATH^G=(k_{ij})_{p\times p}$  would be defined as:
\begin{equation}\label{eq15}
    k_{ij}=
    \begin{cases}
    1 & \text{if there is a path between i \& j}\\
    0 & \text{else}
    \end{cases}
\end{equation}
Using a BFS algorithm and an adjacency matrix, we can check the existence of a path between every two nodes. 
\break \textbf{Lemma}  A sufficient and necessary condition for existing no path between any pair of nodes is:
\begin{equation}\label{eq16}
    \lvert \boldsymbol\beta_{i.j}\times P_{ji}\rvert = 0
\end{equation}
\break \textbf{Proof}  Consider $G$ is a DAG. First, imagine for every $X_i$ and $X_j$ , $\lvert \boldsymbol\beta_{i.j}\times P_{ji}\rvert \neq 0$. Hence, there is a directed edge between $X_i$ and $X_j$, and at least there is a directed path between $X_j$ and $X_i$. It means that there is a directed cycle in graph, which is a contradiction to our presumption that $G$ is a DAG. On the other side, suppose that for every pair of $X_i$ and $X_j$, $\lvert \boldsymbol\beta_{i.j}\times P_{j.i}\rvert = 0$. If $G$ is not a DAG, there is at least one directed cycle in $G$. Hence, there are two nodes $X_i$ and $X_j$ that there is a path from $X_i$ to $X_j$ . This is a contradiction to our presumption that $\lvert \boldsymbol\beta_{i.j}\times P_{ji}\rvert = 0$. Hence, the lemma proved \citep{ref9}.

Regarding the lemma, we can define the following penalty term that guarantees the obtained structure is a DAG.
\begin{equation}\label{eq17}
    DAGPenalty=\lambda_2 \sum_{j\in \textbf{X}/i}^{}{\lvert}{\rvert}\boldsymbol\beta_{i.j}\times P_{ji}{\lvert}{\rvert}_2
\end{equation}
Hence, the score function is:
\begin{equation}\label{eq18}
    \begin{aligned}
     f_{\boldsymbol\lambda}(\boldsymbol\beta) \triangleq  -LL(\boldsymbol\beta)+& \lambda_1\sum_{i=1}^{p}\sum_{j=1}^{p}{\lvert}{\rvert}\boldsymbol\beta_{i.j}{\lvert}{\rvert}\\
    & +\lambda_2\sum_{j\in \textbf{X}/i}^{}{\lvert}{\rvert}\boldsymbol\beta_{i.j}\times P_{ji}{\lvert}{\rvert}_2
    \end{aligned}
\end{equation}
where $\lambda_1$ and $\lambda_2$ are regulation parameters. The greater $\lambda_1$, the more sparse structure and the less accuracy. On the other side, selecting a large $\lambda_1$ , we would have a more sparse structure, but we might miss some edges. 

\section{The Proposed Optimization Algorithm}\label{sec4}

In our algorithm, all the weights are updated at the same time. It means we update $\boldsymbol\beta_{i.j}$ instead of just one component of the vector. In each iteration, a weight vector $\boldsymbol\beta_{i.j}$,  $(i, j) = ([1, ..., P], [1, …, P]), i \neq j$ is selected. Then using SGD method, the score function is updated using equation \ref{eq23} and regarding the selected weight vector.

SGD calculates gradients using just one or a part of the data which are picked randomly instead of all data records. Randomness in SGD causes big variances in estimations which slows down the convergence \citep{ref36}. However, SGD can reach an approximate solution faster than GD. Therefore, SGD has been proposed by many researchers for high-dimensional problems \citep{ref48,ref44, ref47}.

As a common feature of stochastic optimization, SGD in comparison with GD, has a better chance of finding the global optimal solution for complex problems. The deterministic gradient in batch GD may cause the objective function to fall into a local minimum in multimodal problems. The fluctuation in the SGD helps the objective function jump to another possible minimum. However, the fluctuation in SGD always exists, which may result in not achieving a good solution. To tackle this issue, the learning rate $\gamma$ has to decay in each iteration, but smaller $\gamma$ slows down the convergence process. This means that we have a fast computation per iteration but slow convergence for SGD.

To prevent this problem, recently various variance-reduced methods have been proposed \citep{ref36,ref52,ref50}. Using the variance reduction method in SGD, the learning rate for SGD does not have to decay and one can choose a relatively large $\gamma$ which leads to faster convergence \citep{ref51}. Variance reduction techniques cause SGD converges to a more accurate solution faster than regular SGD. Hence, SGD which uses a variance reduction technique is an appropriate method for large-scale problems \citep{ref45,ref46}. We use a variance reduction technique in the optimization step of our proposed algorithm.

To optimize $f_{\boldsymbol\lambda}(\boldsymbol\beta)$ with regard to $\boldsymbol\beta_{i.j}$, $f_{\boldsymbol\lambda,i}(\boldsymbol\beta_{i..})$ would be defined as:
\begin{equation}\label{eq19}
    \begin{aligned}
     f_{\boldsymbol\lambda,i}(\boldsymbol\beta_{i..})= & -\sum_{h\in O_i}^{}\Big[\sum_{\ell =1}^{r_i}y_{hi\ell}\textbf{x} _h^T\boldsymbol\beta_{i\ell.}\\
     & -\log\sum_{m=1}^{r_i}(\textbf{x} _h^T\boldsymbol\beta_{im.} ) \Big] \\
     & +\lambda_1\sum_{j=1}^{p}{\lvert}{\rvert}\boldsymbol\beta_{i.j}{\lvert}{\rvert}_2 \\
     & +\lambda_2\sum_{j\in \boldsymbol X/i}^{}{\lvert}{\rvert}\boldsymbol\beta_{i.j}\times P_{ji}{\lvert}{\rvert}_2
    \end{aligned}
\end{equation}
where $\boldsymbol\beta_{i..}=(\beta_{i.0},\boldsymbol\beta_{i.1},...,\boldsymbol\beta_{i.p}) $. We use an SGD method to optimize equation \ref{eq19}, hence in each iteration, SGD picks a data row randomly and then optimizes \ref{eq19} regarding the selected data. The cost function for data $h$ is:
\begin{equation}\label{eq20}
    \begin{aligned}
         f_{\boldsymbol\lambda,ih}(\boldsymbol\beta_{i..})= & \sum_{\ell =1}^{r_i}y_{hi\ell}\textbf{x}_h^T\boldsymbol\beta_{i\ell .}
    -\log\sum_{m=1}^{r_i}(\textbf{x}_h^T\boldsymbol\beta_{im.})\\
    & +\lambda_1\sum_{j=1}^{p}{\lvert}{\rvert}\boldsymbol\beta_{i.j}{\lvert}{\rvert}_2\\ & +\lambda_2\sum_{j\in \boldsymbol X/i}^{}{\lvert}{\rvert}\boldsymbol\beta_{i.j}\times P_{ji}{\lvert}{\rvert}_2
    \end{aligned}
\end{equation}

Gradient of the cost function is:
\begin{equation}\label{eq21}
    \nabla LL_{ih}(\boldsymbol\beta_{i.j}^{(t)})=\left [ \begin{array}{c}
    (y_{hi1}-p_{i1}^{(t)}(\textbf{x}_h))\textbf{x}_{h,j}\\
     \vdots \\
    (y_{hir_i}-p_{ir_i}^{(t)}(\textbf{x}_h))\textbf{x}_{h,j}
    \end{array}\right]
\end{equation}
To reduce the estimation variance, we have used Stochastic Variance Reduced Gradient (SVRG) \citep{ref36} method. SVRG saves $\hat{\boldsymbol\beta}_{i.j}$ after each $m$ iteration and then calculates $\nabla f_{\boldsymbol\lambda,ih}(\hat{\boldsymbol\beta}_{i.j})$ and $\hat{\mu}$ for $\hat{\boldsymbol\beta}_{i.j}$.
\begin{equation}\label{eq22}
    \hat{\mu}=\frac{1}{n}\sum_{h=1}^{n}\nabla f_{\boldsymbol\lambda,ih}(\hat{\boldsymbol\beta}_{i.j})
\end{equation}
Weight vectors are updated as follow:
\begin{equation}\label{eq23}
    \begin{aligned}
    \boldsymbol\beta_{i.j}^{(t)}=  \boldsymbol\beta_{i.j}^{(t-1)}-\gamma\big(& \nabla f_{\boldsymbol\lambda,ih}(\boldsymbol\beta_{i.j}^{(t-1)})\\
    & -\nabla f_{\boldsymbol\lambda,ih}(\hat{\boldsymbol\beta}_{i.j})+\hat{\mu}\big)
    \end{aligned}
\end{equation}

In equation \ref{eq23} $E[\nabla f_{\boldsymbol\lambda ,i}(\hat{\boldsymbol\beta}_{i.j})-\hat{\mu}]=0 $, hence the expectation of equation \ref{eq23}, which uses variance reduction method, equals to regular SGD.
\begin{equation}\label{eq24}
    E\left[\boldsymbol\beta_{i.j}^{(t)}\mid\boldsymbol\beta_{i.j}^{(t-1)}   \right]=\boldsymbol\beta_{i.j}^{(t-1)}-\gamma\nabla f_{\boldsymbol\lambda ,i}(\boldsymbol\beta_{i.j}^{(t-1)})
\end{equation}

Our proposed algorithm SVRCD is a BCD algorithm that selects a weight vector in each iteration and then optimizes the cost function. The pseudocode is shown in Algorithm \ref{algo2}. SVRCD needs determined regulation parameters $\lambda_1$, $\lambda_2$, and also learning rate $\gamma$.
\begin{algorithm}[t]
\caption{SVRCD Algorithm}\label{algo2}
\begin{algorithmic}
\Require $\gamma>0$, $\lambda>0$, Random $\boldsymbol\beta,  \hat{\boldsymbol\beta}$ 
\Ensure $\boldsymbol\beta$ constructs a BN
\While{stop criterion is not met}
    \For{ $i,j \in \{1, ..., P\}, i \neq j$}
        \For{s=1, 2, ..., S}
            \State $\hat{\boldsymbol\beta}_{i.j}=\hat{\boldsymbol\beta}_{i.j}^{(s-1)}$
            \State $\hat{\mu}_i=\frac{1}{n}\sum_{h=1}^{n}\nabla f_{ih}(\hat{\boldsymbol\beta}_{i.j})$
            \For{t=1, ..., m}
                \State Pick $h \in {1,...,n}$ randomly
                 
                   \State
                    $\boldsymbol\beta_{i.j}^{(t)}=\boldsymbol\beta_{i.j}^{(t-1)}-\gamma\Big(\nabla f_{\boldsymbol\lambda,ih}(\boldsymbol\beta_{i.j}^{(t-1)}$
                    \State \quad \quad \quad $-\nabla f_{\boldsymbol\lambda,ih}(\hat{\boldsymbol\beta}_{i.j})+\hat{\mu}\Big)$
                     
            \EndFor     
 
            \State $\hat{\boldsymbol\beta}_{i.j}^{(s)}=\boldsymbol\beta_{i.j}^{(m)}$
        \EndFor
    \EndFor    
\EndWhile
\State\Return $\boldsymbol\beta$

\end{algorithmic}
\end{algorithm}

\section{Experimental Results}\label{sec5}

\begin{table*}[t]
\caption{Evaluation metrics}
\resizebox{\textwidth}{!}{
\label{tab1}   
\begin{tabular}{lll}
\hline\noalign{\smallskip}
Description & Metric & Abbreviation  \\
\noalign{\smallskip}\hline\noalign{\smallskip}
Number of estimated edges & Predicted & $P$ \\
Number of edges which are in skeleton and have right direction & Expected & $E$ \\
Number of edges which are in skeleton but have wrong direction & Reverse & $R$ \\
Number of edges which are not recognized & Missing & $M=s_0-E-R$ \\
Number of edges which are not in original BN but are in estimated BN & False Positive & $FP=P-E-R$ \\
Right edges learning rate & True Positive Rate & $TPR=\frac{E}{s_0}$ \\
Wrong edges learning rate & False Discovery Rate & $FDR=\frac{R+FP}{P}$ \\
Distance between estimated and original BN & Structural Hamming Distance & $SHD=R+M+FP$ \\
Similarity of estimated and original BN & Jaccard Index & $JI=\frac{E}{P+s_0-E}$\\
\noalign{\smallskip}\hline
\end{tabular}
}
\end{table*}
In this section, we evaluate the proposed algorithm. At first, we tested the algorithm with different parameter settings to find the best values for $\lambda_1$, $\lambda_2$, and $\gamma$. Then we compare SVRCD with other known algorithms such as HC, PC, MMHC, and a new competitive CD algorithm  \citep{ref35}. PC and MMHC are respectively constraint based and hybrid methods proposed for BN structure learning while the other methods are score-based BN structure learning methods. At the end of this section, the algorithm scalability and noise robustness are evaluated.

We have evaluated efficiency, scalability, and robustness of SVRCD using data simulated from three types of graphs: bipartite graph, random graph, and scale-free network. A bipartite graph is one that its vertices can be divided into two disjoint and independent sets $A$ and $B$ in which every edge connects a vertex in $A$ to a vertex in $B$. A random graph is a graph in which the number of vertices, edges and also existence of an edge are determined randomly, for instance using a random distribution. The last type of graphs, scale-free networks, are ones with power-law degree distributions. That is, the fraction $P(k)$ of nodes in the network having k connections to other nodes goes for large values of k as $P(k) ~ k^{-\gamma}$, where $\gamma$ is a parameter whose value is typically in the range $2<\gamma<3$.

To produce bipartite, and scale-free graphs, we have used the igraph package \citep{ref40} in the R environment. Bipartite graphs have $0.2p$ upper and $0.8p$ lower nodes where $p$ is the number of all nodes in the graph, and $s_0=p$ directed edges from upper nodes to lower ones. Scale-free networks are produced using the Barabasi-Albert \citep{ref41} model which has $s_0=p-1$ edges. We have set $t=-3$ and $m=1$, which are two needed parameters for producing scale-free networks in the igraph package. Random graphs have been generated using sparsebnUtils package in R. There are $s_0=p$ edges in our random graphs. 

\begin{table}[t]
\caption{SVRCD evaluation with  changes}
\resizebox{1\columnwidth}{!}{
\label{tab2}   
\begin{tabular}{llllllllll}
\hline\noalign{\smallskip}
$\lambda_1$ & P & E & R & M & FP & TPR & FDR & SHD & JI  \\
\noalign{\smallskip}\hline\noalign{\smallskip}
0.9 & 45 & 26.4 & 9.6 & 14 & 10.4 & 0.53 & 0.43 & 34 & 0.38 \\
1 & 36.8 & 26.6 & 6.4 & 17 & 3.8 & 0.53 & 0.28 & \textbf{27.2} & \textbf{0.44}\\
1.1 & 26.6 & 22 & 3.8 & 24.2 & 0.8 & 0.44 & 0.17 & 28.8 & 0.4 \\
1.2 & 27.6 & 16.8 & 5.2 & 28 & 5.6 & 0.34 & 0.39 & 38.8 & 0.28 \\
1.3 & 22.4 & 14.8 & 3.6 & 31.6 & 4 & 0.3 & 0.34 & 39.2 & 0.26 \\
\noalign{\smallskip}\hline
\end{tabular}
}
\end{table}
We have run HC and MMHC algorithms using bnlearn package \citep{ref42}, and PC algorithm using Pcalg package in R. For CD algorithm, we have used its implementation in discretedAlgorithm package \citep{ref43} in R. We have set CD parameters to their default values. All variables in graphs are considered to be binary in all implementations. Weight vectors $\boldsymbol\beta_{i.j}=(\beta_{i1j}, \beta_{i2j})$  have been initialized randomly with a value in range $(0,1)$. 
We have considered different values for  $(n,p)$ in evaluations, in which $n$ refers to the number of data rows and $p$ indicates the number of variables of the problem. The goal is to evaluate SVRCD using high-dimensional discrete data; hence we have set $p \geq n$ in data generation process. Different evaluation metrics are used in this paper that are introduced in table \ref{tab1}. All results are averages over 20 datasets for each $(n,p)$ setting.

\subsection{Efficiency}
In the first step of evaluations, we measure the impact of $\lambda_1$ on evaluation metrics. We have reported the results in Table \ref{tab2}. As shown in the table, in $\lambda_1=0.9$, $P$ has the greatest value, since $\lambda_1$ determines how sparse the structure can be. We should set $\lambda_1$ to a proper value that controls sparsity but does not decrease other metrics such as $E$. SVRCD has the best SHD and JI in $\lambda_1=1$ for $(n,p)=(50,50)$. In addition, our proposed algorithm has a greater P in $\lambda_1=0.9$, but a greater FP, FDR, SHD, and a lower JI as well. Also, SVRCD estimates a sparser structure using $\lambda_1=1.3$ but E has a low value in this setting.

\begin{table}[t]
\caption{SVRCD evaluation with $\lambda_2$ changes}
\resizebox{1\columnwidth}{!}{
\label{tab3}   
\begin{tabular}{llllllllll}
\hline\noalign{\smallskip}
$\lambda_2$ & P & E & R & M & FP & TPR & FDR & SHD & JI  \\
\noalign{\smallskip}\hline\noalign{\smallskip}
0.1 & 38.8 & 24.2 & 9 & 16.8 & 5.6 & 0.48 & 0.37 & 31.4 & 0.37 \\
0.2 & 36.8 & 26.6 & 6.4 & 17 & 3.8 & 0.53 & 0.28 & \textbf{27.2} & \textbf{0.44}\\
0.3 & 32.6 & 24.2 & 7.4 & 18.4 & 3.8 & 0.48 & 0.31 & 29.6 & 0.4 \\
0.4 & 31.4 & 21.2 & 6.6 & 22.2 & 3.6 & 0.42 & 0.33 & 32.4 & 0.35 \\
0.5 & 31.4 & 18.2 & 8.8 & 23 & 4.4 & 0.36 & 0.42 & 36.2 & 0.29 \\
\noalign{\smallskip}\hline
\end{tabular}
}
\end{table}

Table \ref{tab3} shows the results for different values of $\lambda_2$. This parameter puts a constraint on the structure to be a DAG. Also, $\lambda_2$ has a small effect on sparsity, hence selecting a proper value for this parameter is crucial. A large value of $\lambda_2$ causes a large FP and a low TP, and a low value of $\lambda_2$ causes the structure not to be a DAG. The results show that with our generated dataset, the best value for this parameter is 0.2. Although SVRCD has the lowest FP in $\lambda_2=0.4$, E, TPR, FDR, SHD, and JI have the best values in $\lambda_2=0.2$. Hence one will observe a better overall performance in $\lambda_2=0.2$.

\begin{table}[t]
\caption{SVRCD evaluation with $\gamma$ changes}
\resizebox{1\columnwidth}{!}{
\label{tab4}   
\begin{tabular}{llllllllll}
\hline\noalign{\smallskip}
$\gamma$ & P & E & R & M & FP & TPR & FDR & SHD & JI  \\
\noalign{\smallskip}\hline\noalign{\smallskip}
0.001 & 36.8 & 26.6 & 6.4 & 17 & 3.8 & 0.53 & 0.53 & \textbf{27.2} & \textbf{0.44} \\
0.002 & 38.2 & 24.2 & 8.6 & 17.2 & 5.4 & 0.48 & 0.48 & 31.2 & 0.38 \\
0.004 & 40.8 & 25.8 & 9.4 & 14.8 & 5.6 & 0.52 & 0.52 & 29.8 & 0.4 \\
0.006 & 43.4 & 27 & 7.4 & 15.6 & 9 & 0.54 & 0.54 & 32 & 0.41 \\
0.008 & 49 & 24 & 12 & 14 & 13 & 0.48 & 0.48 & 39 & 0.32 \\
\noalign{\smallskip}\hline
\end{tabular}
}
\end{table}
\begin{table*}[t]
\caption{Comparison among SVRCD and other algorithms on bipartite graph, random DAG, and scale-free network data}

\label{tab5}   
\resizebox{\textwidth}{!}{
\begin{tabular}{lllllllllll}
\hline\noalign{\smallskip}
Graph & Method & P & E & R & M & FP & TPR & FDR & SHD & JI  \\
\noalign{\smallskip}\hline\noalign{\smallskip}
Bipartite & SVRCD & 105 & 73 & 16.67 & 111.33 & 16.33 & 0.37 & 0.29 & \textbf{143.33} & \textbf{0.32} \\
 & BCD & 84 & 51.95 & 18.65 & 129.4 & 13.4 & 0.26 & 0.37 & 161.45 & 0.22 \\
 & PC & 75.7 & 26.9 & 34.2 & 138.9 & 14.6 & 0.13 & 0.64 & 187.7 & 0.18 \\
 & HC & 378.1 & 111.5 & 32.9 & 55.6 & 233.8 & 0.56 & 0.71 & 322.4 & 0.24 \\
 & MMHC & 175.4 & 72.2 & 20.4 & 107.4 & 82.8 & 0.36 & 0.59 & 210.1 & 0.24 \\
 & NoCurl & 3686 & 29.8 & 38 & 128 & 1776 & 0.15 & 0.98 & 1942 & 0.01 \\

Random DAG & SVRCD & 95.5 & 67.95 & 18.3 & 113.7 & 9.45 & 0.34 & 0.29 & \textbf{141.50} & \textbf{0.30} \\
 & BCD & 85.75 & 56 & 18.6 & 125.4 & 11.15 & 0.28 & 0.34 & 155.15 & 0.24 \\
 & PC & 97.3 & 47.1 & 37.0 & 138.9 & 13.2 & 0.23 & 0.52 & 169.7 & 0.19 \\
 & HC & 376.3 & 96.3 & 52.1 & 55.6 & 227.8 & 0.48 & 0.74 & 335.1 & 0.20 \\
 & MMHC & 179.8 & 86.4 & 31.1 & 107.4 & 62.4 & 0.43 & 0.52 & 179.6 & 0.29 \\
 & NoCurl & 3745 & 32 & 41 & 134 & 1695 & 0.17 & 0.95 & 1928 & 0.01 \\
 
Scale-free & SVRCD & 73.35 & 50.25 & 16.70 & 132.05 & 6.40 & 0.25 & 0.31 & \textbf{155.15} & 0.23 \\
 & BCD & 80.10 & 45.80 & 23.45 & 129.75 & 10.85 & 0.23 & 0.42 & 164.05 & 0.20 \\
 & PC & 99.50 & 46.50 & 23.10 & 138.90 & 30.00 & 0.23 & 0.53 & 182.50 & 0.19 \\
 & HC & 377.80 & 121.00 & .1028 & 55.6 & 228.80 & 0.61 & 0.68 & 306.90 & 0.27 \\
 & MMHC & 176.80 & 93.20 & 16.10 & 107.40 & 67.50 & 0.47 & 0.47 & 173.20 & \textbf{0.33}\\
 & NoCurl & 3711 & 35 & 39 & 132 & 1764 & 0.15 & 0.96 & 1935 & 0.01 \\
 
\noalign{\smallskip}\hline
\end{tabular}
}
\end{table*}

Table \ref{tab4} reports measurement metrics with regard to different values for learning rate $\gamma$. A large value for $\gamma$ results in a wrong learnt structure and a large FDR and FP. Furthermore, a low $\gamma$ increases the runtime. We have found $\gamma=0.001$ the best setting in our evaluations. SVRCD has a better TPR in $\gamma=0.006$, but a larger SHD. SHD and JI specify the quality of the estimated structure. Hence, $\gamma=0.001$ is the best setting in our evaluations.

\begin{table}[h]
\caption{Scalability of SVRCD on bipartite graph data}
\resizebox{1\columnwidth}{!}{
\label{tab8}   
\begin{tabular}{llllllllll}
\hline\noalign{\smallskip}
(n,p) & P & E & R & M & FP & TPR & FDR & SHD & JI  \\
\noalign{\smallskip}\hline\noalign{\smallskip}
$(50,50)$ & 36.2 & 25 & 4.8 & 20.2 & 6.4 & 0.5 & 0.3 & 31.4 & 0.41 \\
$(50,100$ & 53.6 & 37.4 & 7.6 & 55 & 8.6 & 0.37 & 0.3 & 71.2 & 0.32 \\
$(100,100)$ & 65.2 & 45.2 & 13.2 & 41.6 & 6.8 & 0.45 & 0.31 & 61.6 & 0.38 \\
$(50,200)$ & 105 & 73 & 16.67 & 111.33 & 16.33 & 0.37 & 0.29 & 143.33 & 0.32 \\
\noalign{\smallskip}\hline
\end{tabular}}
\end{table}

The second step of evaluations compares SVRCD with some known algorithms. First, the datasets are sampled from a bipartite graph with $(n,p)=(50,200)$ setting. As shown in Table \ref{tab5}, our proposed algorithm outperforms BCD, PC, HC, and MMHC algorithms regarding FDR, SHD, and JI metrics. Although HC demonstrates better TPR and E than SVRCD, it estimates 387.1 edges for a BN with 200 edges. Hence, the estimated structure is not sparse and has more edges than the original structure. SVRCD estimated two more FP edges than the PC algorithm. The reason is that the PC has estimated fewer edges, and has a lower E. SVRCD with $E=73$ has defeated the PC with $E = 26.9$. In addition, SVRCD has better FDR, TPR, SHD, and JI than PC. Hence, SVRCD generally outperformed PCs. BCD is one of the recently proposed algorithms for estimating sparse BN using high-dimensional discrete data. We compared SVRCD to this algorithm as well. Our proposed algorithm leads to much better SHD, JI, FDR, TPR, and E than BCD, but BCD has estimated a sparser structure and has a lower FP. However, SHD and JI are two important metrics that show the overall performance of a structure learning algorithm.

We also compared SVRCD with a recent structure learning method called NoCurl \citep{ref32} which originally has been proposed for continuous data. Although NoCurl has competitive results in the continuous domain, it does not achieve good results in learning DAGs using discrete data, and SVRCD outperforms it in all metrics. Furthermore, we have repeated our evaluations with data sampled from random graphs and scale-free networks. The results reported in Table \ref{tab5} shows that SVRCD demonstrates the best SHD and JI in learning Random DAGs. But, for scale-free networks, although it achieves SHD=155.15, which is the best level among all six algorithms, MMHC with JI=0.33 outperforms SVRCD.

\subsection{Scalability}
The next set of evaluations examines the scalability of SVRCD with respect to $(n , p)$. The results have been reported in Table \ref{tab8}. The greater $p$ and the lower $n$ are, the more sophisticated problem is. We obtain from Table \ref{tab8} that SVRCD is more efficient for $(n,p)=(100,100)$ than $(n,p)=(50,100)$. On the other hand, comparing $(n,p)=(50,100)$ to $(n,p)=(50,200)$, we observe that SVRCD achieves the same TPR, FDR, and JI. It means that increasing the number of variables does not have a significant negative effect on the TPR, FDR, and JI. SVRCD can obtain admissible results on high-dimensional data.

\begin{figure*}
\centering
  \includegraphics[width=0.75\textwidth]{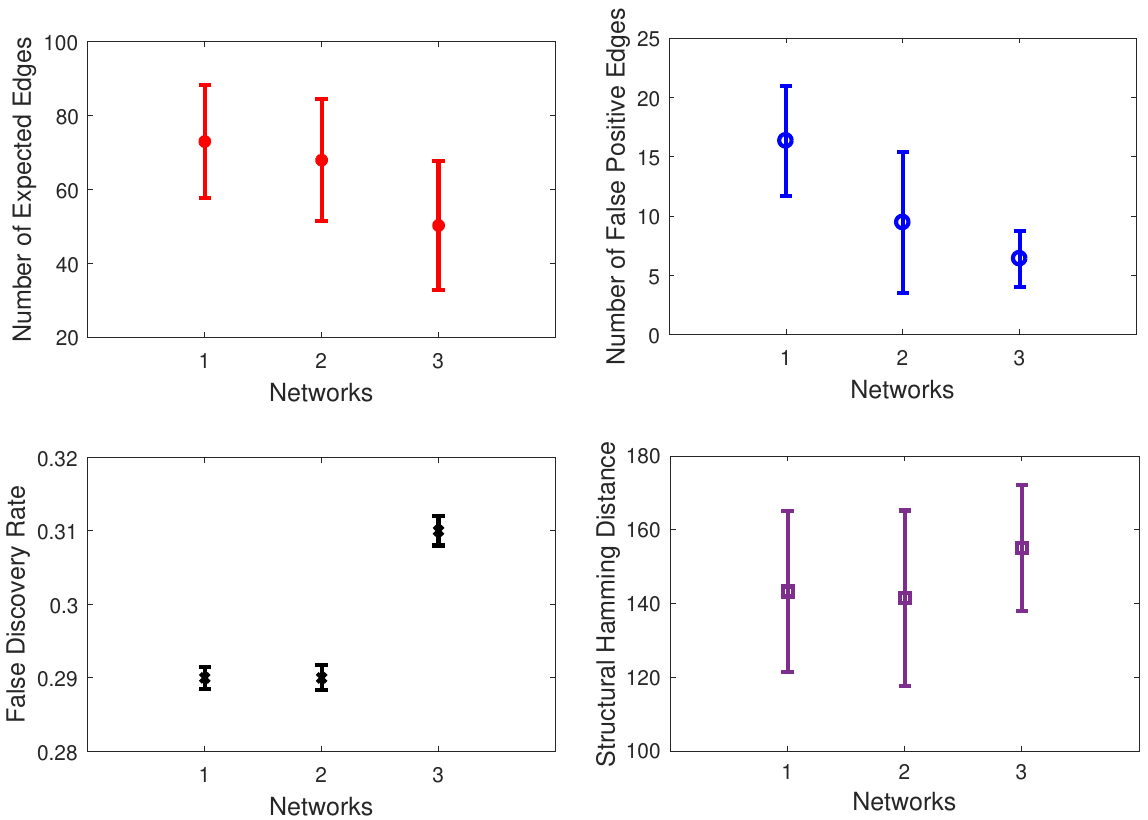}
  \caption{Means and variances of E, FP, FDR, and SHD in estimating bipartite graph, random graph, and scale-free net.}
\label{fig1}       
\end{figure*}

\begin{figure*}
\centering
  \includegraphics[width=0.7\textwidth]{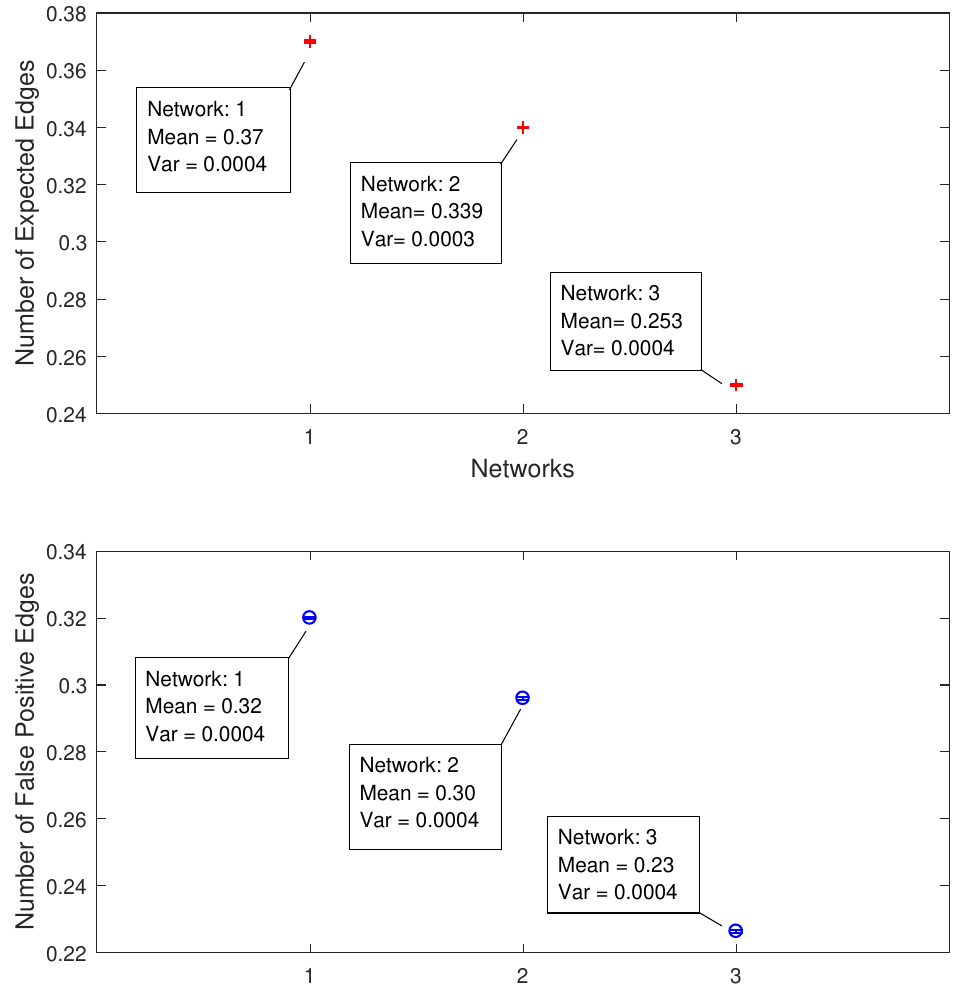}
\caption{Means and variances of TPR, and JI in estimating bipartite graph, random graph, and scale-free net.}
\label{fig2}       
\end{figure*}
\begin{figure*}[t]
\centering
  \includegraphics[width=0.7\textwidth]{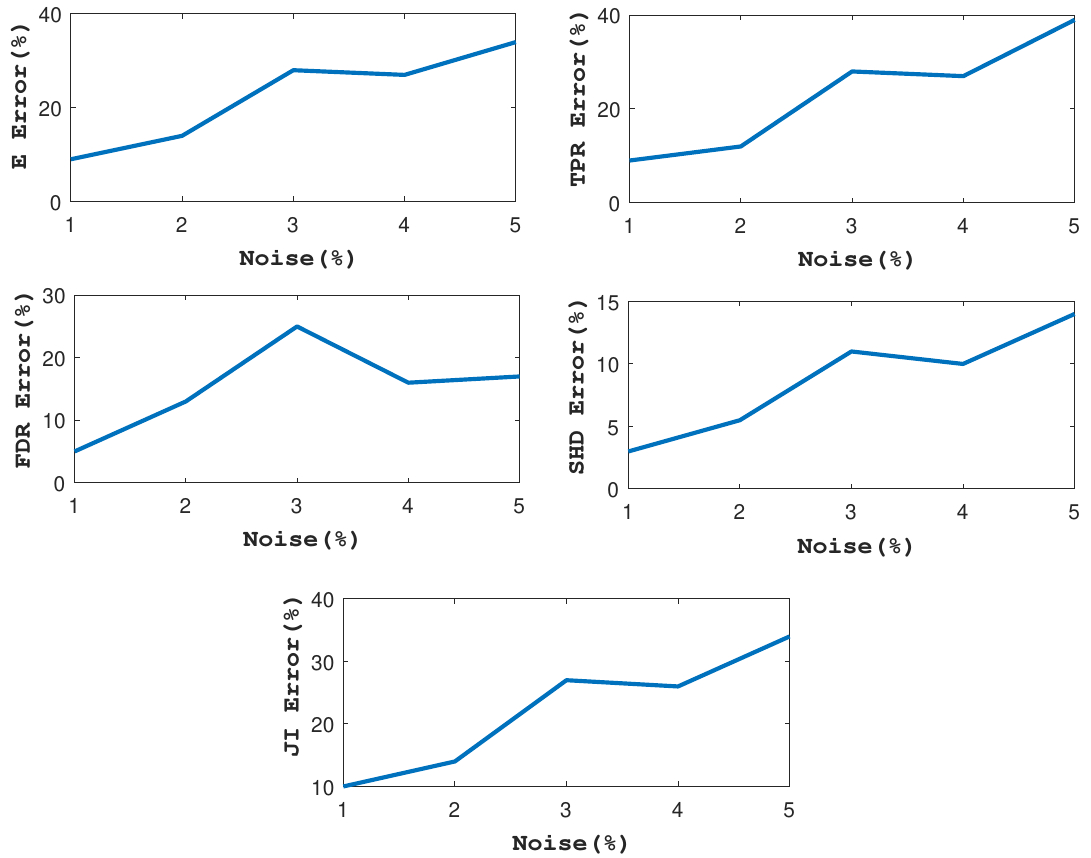}
\caption{E, TPR, FDR, SHD, and JI changes with different amount of noise in estimating bipartite graph.}
\label{fig3}       
\end{figure*}

\begin{table}[t]
\caption{SVRCD robustness against noise on bipartite graph data}
\resizebox{1\columnwidth}{!}{
\label{tab9}   
\begin{tabular}{llllllllll}
\hline\noalign{\smallskip}
Noise\% & P & E & R & M & FP & TPR & FDR & SHD & JI  \\
\noalign{\smallskip}\hline\noalign{\smallskip}
0 & 86.4 & 64 & 14.2 & 121.8 & 8.2 & 0.32 & 0.26 & 144.2 & 0.29 \\
1 & 79.2 & 58.2 & 14 & 127.8 & 7 & 0.29 & 0.26 & 148.8 & 0.26 \\
2 & 72.6 & 55 & 10.2 & 134.8 & 7.4 & 0.28 & 0.24 & 152.4 & 0.25 \\
3 & 66.8 & 46 & 14.8 & 139.2 & 6 & 0.23 & 0.31 & 160 & 0.21 \\
4 & 64.6 & 46 & 13.6 & 140.4 & 5 & 0.23 & 0.29 & 159 & 0.21 \\
5 & 58.8 & 42 & 10.6 & 147.4 & 6.2 & 0.21 & 0.29 & 164.2 & 0.19\\
\noalign{\smallskip}\hline
\end{tabular}}
\end{table}

In the fourth step of evaluations, we have tested SVRCD using datasets with $(n,p)=(50,200)$ settings generated from bipartite graphs, random graphs, and scale-free networks to measure the performance of our proposed algorithm in estimating each graph. For each test, we generated 20 datasets from each graph. Figure \ref{fig1} represents the means and variances of E, FP, FDR, and SHD while estimating each graph using 20 datasets. We have reported JI, and TPR means and variances in Figure \ref{fig2} with more details since the variances are too small. 
In the horizontal axis, 1, 2, and 3 represent the bipartite graph, random graph, and scale-free network, respectively.

\subsection{Robustness}
The final part of evaluations calculates the algorithm robustness against noise. We have sampled data from a bipartite graph with $(n,p)=(50,200)$. Then we have swapped a specific portion of data to their complement values. The results are reported in Table \ref{tab9} and Figure \ref{fig3}. From Figure \ref{fig3}, we observe that increasing the noise causes an increase in E and TPR errors. But, FDR increases at first and then decreases, since increasing the noise causes the algorithm to predict fewer edges (smaller P).

\section{Conclusion}\label{sec6}

In this paper, we proposed a novel BN structure learning algorithm, SVRCD, for learning BN structures from discrete high-dimensional data. Also, we proposed a score function consisting of three parts: likelihood, sparsity penalty term, and DAG penalty term. SVRCD is a BCD algorithm that uses a variance reduction method in the optimization step, which is an SGD method. Variance reduction method is used to prevent large variances in estimations and hence, increases the convergence speed. As shown in the previous section, our algorithm outperforms some known algorithms in BN structure learning. SVRCD shows significant results in learning BN structure from discrete high-dimensional data. One of the advantages of SVRCD is that more data rows causes a lower algorithm runtime. Because it uses an SGD method and when the data rows increase, the algorithm iterations should be set on a lower value to prevent the algorithm overfitting.  Obtained results show that SVRCD dominates the recently proposed BCD algorithm in SHD and JI metrics. 

In future work, we will investigate and prove the algorithm convergence for different discrete data distributions. Furthermore, this algorithm is proposed for learning discrete BN, but the same formulation may be adopted for Gaussian BN. In addition, since BNs have many applications in medical studies, SVRCD may help medical problems such as constructing gene networks. Another potential future work can be finding tuning parameters causing the algorithm to converge to the optimum point.

\section*{Declarations}

\begin{itemize}
\item Funding: No funds, grants, or other support was received.
\item Conflict of interest/Competing interests: The authors have no conflicts of interest to declare.
\item Availability of data and materials: The data generation method is available online\footnote{\label{note1}\url{https://github.com/nshajoon/SVRCD-Algorithm.git}}.
\item Code availability: The code is available online\repeatfoot.

\end{itemize}

\bibliography{sn-bibliography}

\end{document}